\documentclass[letterpaper, 10 pt, conference]{ieeeconf}  
\usepackage{amsmath}
\usepackage{amssymb}
\usepackage{booktabs}
\usepackage{caption}
\usepackage{comment}
\usepackage{graphicx}
\usepackage{multirow}
\usepackage{pifont}
\usepackage[table]{xcolor}
\definecolor{darkgreen}{RGB}{0, 150, 0}
\definecolor{darkred}{RGB}{200, 0, 0}

\makeatletter
\let\NAT@parse\undefined
\makeatother

\newcommand{\ie}{i.e.}
\newcommand{\ch}{{\color{darkgreen} \ding{51}}}
\newcommand{\xm}{{\color{darkred} \ding{55}}}
\usepackage[colorlinks,pagebackref=true,citecolor=green,bookmarks=false,hypertexnames=true]{hyperref}

\title{\LARGE \bf
FisheyeDistanceNet: Self-Supervised Scale-Aware Distance Estimation using Monocular Fisheye Camera for Autonomous Driving
}

\author{
Varun Ravi Kumar$^{1}$,
Sandesh Athni Hiremath$^{1}$,
Markus Bach$^{1}$,
Stefan Milz$^{1}$, \\
Christian Witt$^{1}$,
Clément Pinard$^{2}$, Senthil Yogamani$^{3}$ and Patrick M\"ader$^{4}$ \\ 
$^{1}$Valeo DAR Kronach, Germany \hspace{0.3cm}
$^{2}$ENSTA ParisTech Palaiseau, France \\
$^{3}$Valeo Vision Systems, Ireland \hspace{0.3cm}
$^{4}$Technische Universit\"at Ilmenau, Germany
}

\begin{document}
\maketitle
\thispagestyle{empty}
\pagestyle{empty}
\begin{abstract}

Fisheye cameras are commonly used in applications like autonomous driving and surveillance to provide a large field of view ($>180^{\circ}$). However, they come at the cost of strong non-linear distortions that require more complex algorithms. In this paper, we explore Euclidean distance estimation on fisheye cameras for automotive scenes. Obtaining accurate and dense depth supervision is difficult in practice, but self-supervised learning approaches show promising results and could overcome the problem. We present a novel self-supervised scale-aware framework for learning Euclidean distance and ego-motion from raw monocular fisheye videos without applying rectification. While it is possible to perform a piece-wise linear approximation of fisheye projection surface and apply standard rectilinear models, it has its own set of issues like re-sampling distortion and discontinuities in transition regions. To encourage further research in this area, we will release our dataset as part of the WoodScape project \cite{yogamani2019woodscape}. We further evaluated the proposed algorithm on the KITTI dataset and obtained state-of-the-art results comparable to other self-supervised monocular methods. Qualitative results on an unseen fisheye video demonstrate impressive performance\footnote{see Fig.~\ref{fig:overview} and \url{https://youtu.be/Sgq1WzoOmXg}}.
\end{abstract}
 
\section{\textbf{Introduction}}

There has been a significant rise in the usage of fisheye cameras in various automotive applications~\cite{heimberger2017computer, dahal2019deeptrailerassist, horgan2015vision},  surveillance~\cite{drulea2014omnidirectional} and robotics~\cite{caruso2015large} due to their large Field of View (FOV). Recently, several computer vision tasks on fisheye cameras have been explored including object detection~\cite{sistu2019real}, soiling detection~\cite{uvrivcavr2019soilingnet}, motion estimation~\cite{yahiaoui2019fisheyemodnet}, image restoration~\cite{uricar2019desoiling} and SLAM~\cite{tripathi2020trained}. 
Depth estimation is an essential task in autonomous driving as it is used to avoid obstacles and plan trajectories. While depth estimation has been substantially studied for narrow FOV cameras, it has barely been explored for fisheye cameras~\cite{varun18,zioulis2018omnidepth}.

Previous learning-based approaches~\cite{godard2019digging, zhou2017unsupervised, monodepth17, Wang_2018_CVPR} have solely focused on traditional 2D content captured with cameras following a typical pinhole projection model based on rectified image sequences.
With the surge of efficient and cheap wide-angle fisheye cameras and their larger FOV in contrast to pinhole cameras, there has been significant interest in the computer vision community to perform depth estimation from omnidirectional content similar to traditional 2D content via omnidirectional stereo~\cite{li2008binocular, ma20153d, pathak2016dense, li2005spherical} and structure-from-motion (SfM)~\cite{huang20176} approaches.

\begin{figure}[!t]
  \captionsetup{singlelinecheck=false, font=small, labelsep=space, belowskip=-2pt}
  \centering
  \newcommand{\turnwidth}{0.485\columnwidth}

\newcommand{\imlabel}[2]{\includegraphics[width=0.49\columnwidth]{#1}%
\raisebox{2pt}{\makebox[-2pt][r]{\footnotesize #2}}}

\begin{tabular}{@{\hskip 0mm}c@{\hskip 1.5mm}c}
\centering
\imlabel{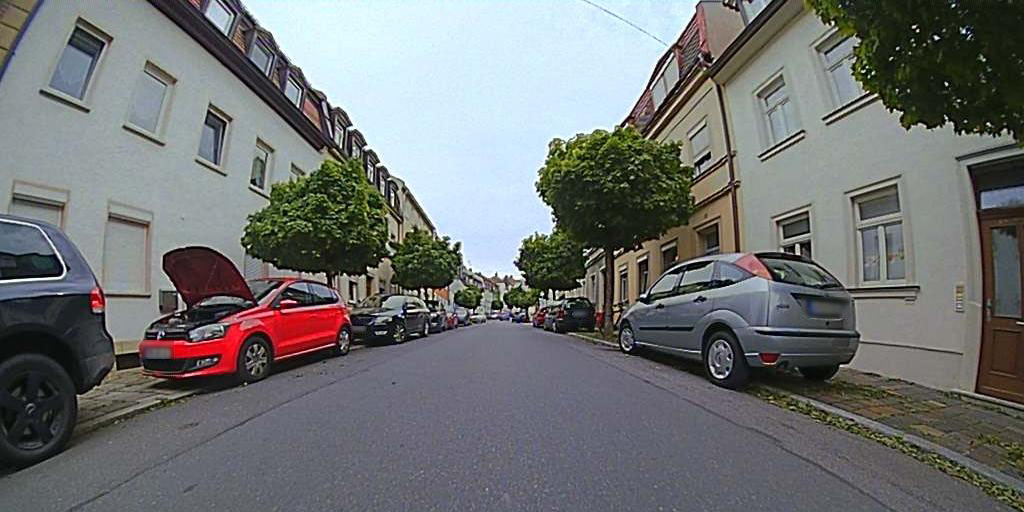}
{\textcolor{white}{WoodScape}} &
\imlabel{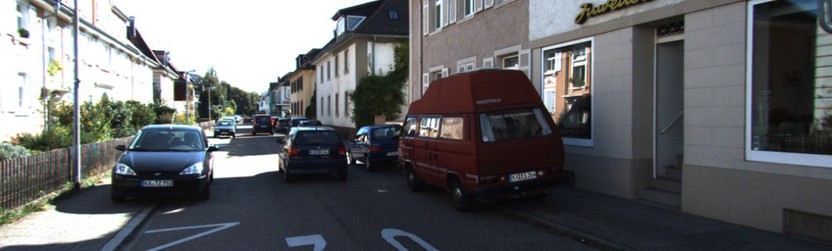}
{\textcolor{white}{KITTI}} \\

\imlabel{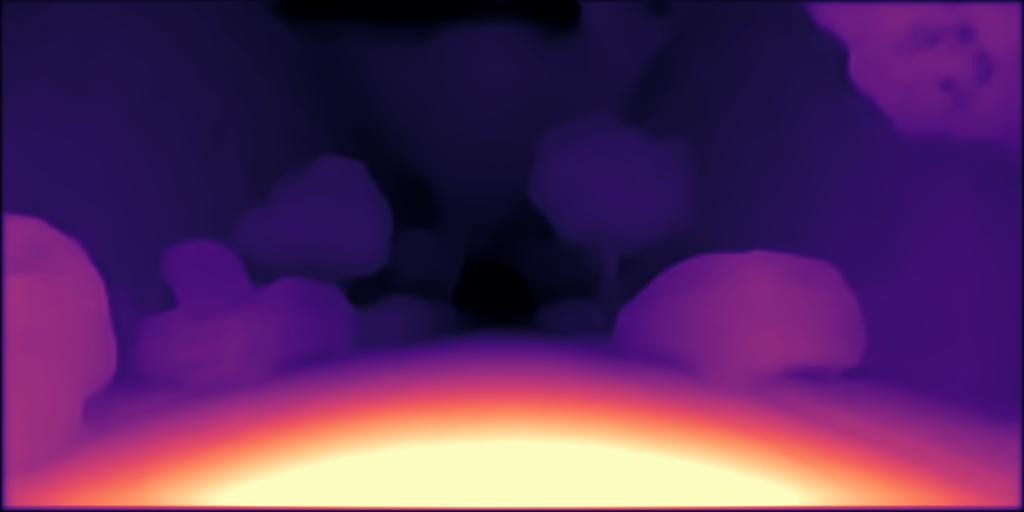}
{} &
\imlabel{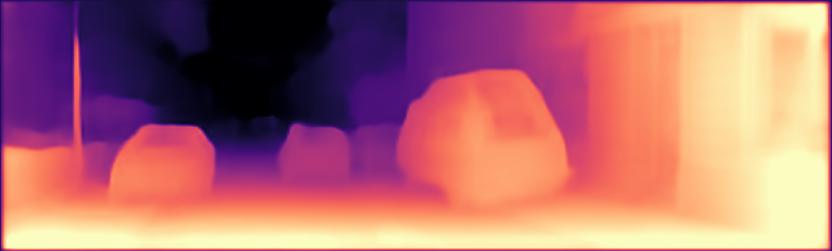}
{} \\
\end{tabular}

  \vspace{0pt}
  \caption{{\bf Distance and depth derived from a single fisheye image (left) and single pinhole image (right) respectively.} Our self-supervised model, \textbf{\textit{FisheyeDistanceNet}}, produces sharp, high quality distance and depth maps.
  }
  \vspace{-14pt}
  \label{fig:overview}
\end{figure}


Depth estimation models may be learned in a supervised fashion on LiDAR distance measurements, such as KITTI~\cite{geiger2013vision}. In previous work, we followed this approach and demonstrated the possibility to estimate high-quality distance maps using LiDAR ground truth on fisheye images~\cite{varun18}. However, setting up the entire rig for such recordings is expensive and time-consuming, limiting the amount of data on which a model can be trained.
\newcommand{\cmt}[1]{\textcolor{blue}{[sandesh] #1}}
To overcome this problem, we propose {\em FisheyeDistanceNet} the first end-to-end self-supervised monocular scale-aware training framework. FisheyeDistanceNet uses convolutional neural networks (CNN) on raw fisheye image sequences to regress a Euclidean distance map and provides a baseline for single frame Euclidean distance estimation. We summarize our contributions  as follows:
\begin{itemize}
\item A self-supervised training strategy aims to infer a distance map from a sequence of distorted and unrectified raw fisheye images.
\item A solution to the scale factor uncertainty with the bolster from ego-motion velocity allows outputting metric distance maps. This facilitates the map's practical use for self-driving cars.
\item A novel combination of super-resolution networks and deformable convolution layers~\cite{zhu2019deformable} to output high-resolution distance maps with sharp boundaries from a low-resolution input. Inspired by the super-resolution of images approach~\cite{shi2016real}, this approach allows us to accurately resolve distances replacing the deconvolution~\cite{odena2016deconvolution} and a naive nearest neighbor or bilinear upsampling.
\item We depict the importance of using backward sequences for training and construct a loss for these sequences. Moreover, a combination of filtering static pixels and an ego mask is employed. The incorporated bundle-adjustment framework~\cite{zhou2018unsupervised} jointly optimizes distances and camera poses within a sequence by increasing the baseline and providing additional consistency constraints. 
\end{itemize}
\section{\textbf{Self-Supervised Scale-Aware FisheyeDistanceNet}}

Zhou et al.'s~\cite{zhou2017unsupervised} self-supervised monocular structure-from-motion (SfM) framework aims at learning:
\begin{enumerate}
\item a monocular depth model $g_d: I_t \to D$ predicting a scale-ambiguous depth $\hat D = g_d(I_t(p))$ per pixel $p$ in the target image $I_t$; and
\item an ego-motion predictor $g_x: (I_t, I_{t^\prime}) \to I_{t \to t^\prime}$ predicting a set of six degrees of freedom rigid transformations $T_{t \rightarrow t'} \in \text{SE(3)}$, between the target image $I_t$ and the set of reference images $I_{t^\prime}$.
Typically, $t' \in \{t+1, t-1\}$, \ie the frames $I_{t-1}$ and $I_{t+1}$ are used as reference images, although using a larger window is possible.
\end{enumerate}
A limitation of this approach is that both depth and pose are estimated up to an unknown scale factor in the monocular SfM pipeline.

\begin{figure}[t]
  \captionsetup{singlelinecheck=false, font=small, labelsep=space, belowskip=-2pt}
  \centering
    \includegraphics[width=\columnwidth]{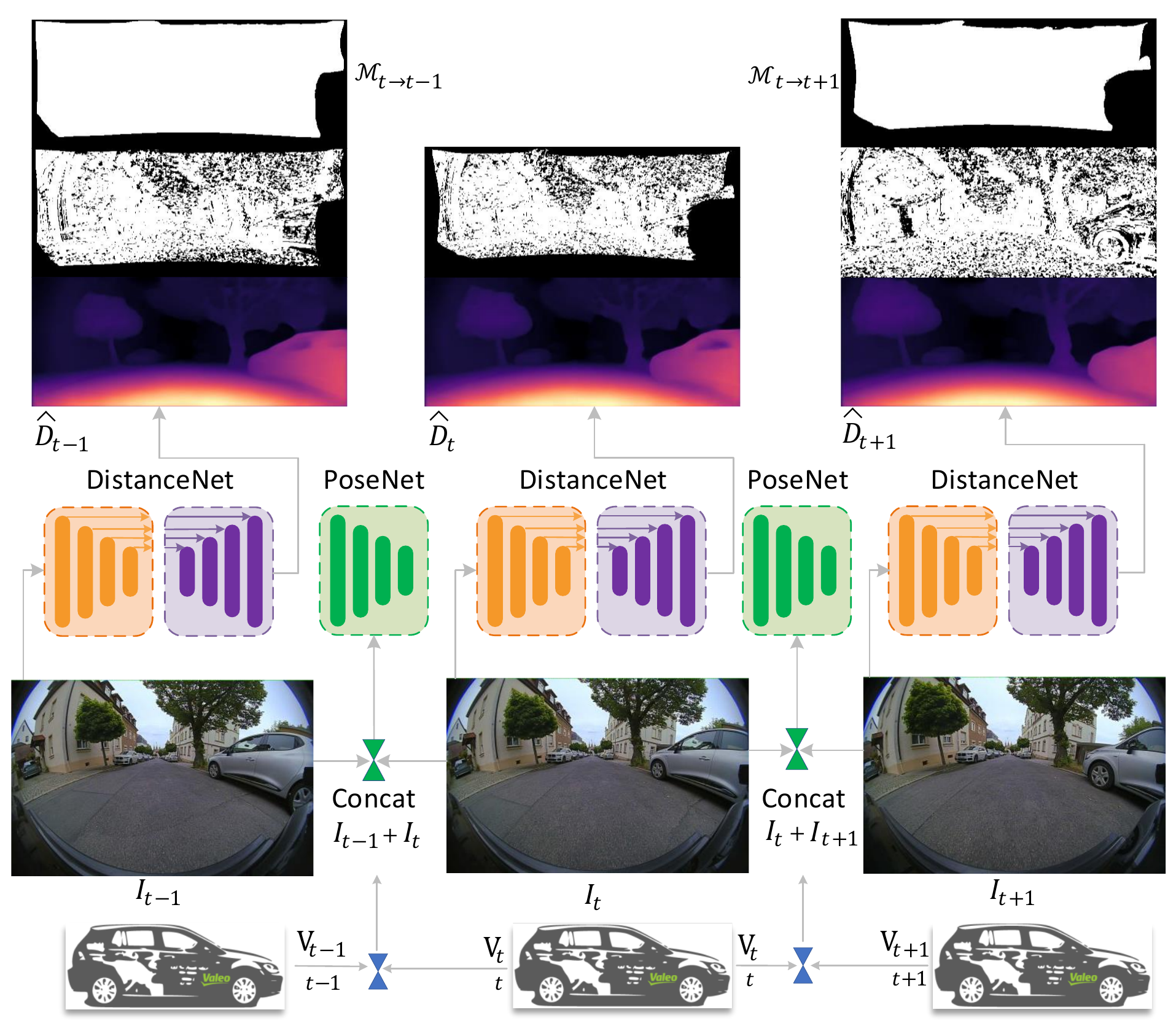}
    \caption{\textbf{Overview of our method.}
    The first row represents our ego masks as described in Section~\ref{sec:mask}, $\mathcal{M}_{t \to t-1}$, $\mathcal{M}_{t \to t+1}$ indicate which pixel coordinates are valid when constructing $\hat I_{t-1 \to t}$ from $I_{t-1}$ and $\hat I_{t+1 \to t}$ from $I_{t+1}$ respectively. The second row indicates the masking of static pixels computed after 2 epochs, where black pixels are filtered from the photometric loss (\ie $\omega=0$). It prevents dynamic objects at similar speed as the ego car and low texture regions from contaminating the loss. The masks are computed for forward and backward sequences from the input sequence $S$ and reconstructed images using Eq. \ref{eqn:staticmask} as described in Section \ref{sec:mask}. The third row represents the distance estimates corresponding to their input frames. Finally, the vehicle's odometry data is used to resolve the scale factor issue.}
    \label{fig:model_arch}
\end{figure}

The depth which acts as an intermediary variable is obtained from the network by constraining the model to perform image synthesis. Depth estimation is an ill-posed problem as there could exist a large number of possible incorrect depths per pixel, which can also recreate the novel view, given the relative pose between $I_t$ and $I_{t^\prime}$.

Using view-synthesis as the supervising technique we can train the network using the viewpoint of $I_{t-1}$ and $I_{t+1}$ to estimate the appearance of a target image $I_t$ on raw fisheye images.
A naive approach would be correcting raw fisheye images to piecewise or cylindrical projections and would essentially render the problem equivalent to Zhou et al.'s work~\cite{zhou2017unsupervised}. In contrast, at the core of our approach there is a simple yet efficient technique for obtaining scale-aware distance maps.

This section starts with discussing the geometry of the problem and how it is used to obtain differentiable losses. We describe the scale-aware FisheyeDistanceNet and its effects on the output distance estimates. Additionally, we provide an in-depth discussion of the various losses.

\newcommand{\iPi}{\Pi^{-1}}
\subsection{\textbf{\textit{Modeling of Fisheye Geometry}}}
\label{sec:modeling of fisheye geometry}

\textbf{\textit{Projection from camera coordinates to image coordinates}}
\label{sec:cam2img}
The projection function $X_c \mapsto \Pi(X_c) = p$ of a 3D point $X_c = (x_c, y_c, z_c)^T$ in camera coordinates to a pixel $p = (u, v)^T$ in the image coordinates is obtained via a $4^\text{th}$ order polynomial in the following way:
\begin{align}
        \varphi &= \text{arctan2}(y_{c},x_{c})\\
    \theta &= \frac{\pi}{2} - \text{arctan2} ({z_c},{r_c}) \\
    \varrho(\theta) &= k_1 \cdot \theta + k_2 \cdot \theta^2 + k_3 \cdot \theta^3 + k_4 \cdot \theta^4 \label{eq:poly4}\\
    p &= \begin{pmatrix} u \\ v \end{pmatrix} = \begin{pmatrix} \varrho(\theta) \cdot \cos\varphi \cdot a_x + c_x \\  \varrho(\theta) \cdot \sin\varphi \cdot a_y + c_y \end{pmatrix} 
\end{align}
where $r_c = \sqrt{x_c^2 + y_c^2}$, $\theta$ is the angle of incidence, $\varrho(\theta)$ is the mapping of incident angle to image radius, $(a_x, a_y)$ is the aspect ratio and $(c_x, c_y)$ is the principal point.\\
\textbf{\textit{Unprojection from image coordinates to camera coordinates}}
\label{sec:img2cam}
The unprojection function $(p,\hat D) \mapsto \iPi(p,\hat D) = X_c$ of an image pixel $p =  (u, v)^T$ and it's distance estimate $\hat D$ to the 3D point $X_c = (x_c, y_c, z_c)^T$ is obtained via the following steps. Letting $(x_i, y_i)^T = \big((u - c_x)/a_x, (v - c_y)/a_y \big)^T$, we obtain the angle of incidence $\theta$ by numerically calculating the $4^\text{th}$ order polynomial roots of $\varrho = \sqrt{x_i^2 + y_i^2}$ using the distortion coefficients $k_1, k_2, k_3, k_4$ (see Eq. \ref{eq:poly4}). For training efficiency, we pre-calculate the roots and store them in a lookup table for all the pixel coordinates. Now, $\theta$ is used to get
\begin{equation}
    r_c = \hat D \cdot \sin\theta \quad \text{ and } \quad z_c = \hat D \cdot \cos\theta
\end{equation}
where the distance estimate $\hat D$ from the network represents the Euclidean distance $\| X_c \|$ =  $\sqrt{x_c^2 + y_c^2 + z_c^2}$ of a 3D point $X_c$. The polar angle $\varphi$ and the $x_c$, $y_c$ components can be obtained as follows:
\begin{equation*}
    \begin{array}{l l l}
           \varphi = \text{arctan2} (y_i, x_i),& \hspace{-.1cm} x_c = r_c \cdot \cos\varphi,& \hspace{-.1cm} y_c = r_c \cdot \sin\varphi.
    \end{array}
\end{equation*}
\subsection{\textbf{\textit{Photometric Loss}}}
\label{sec:photometric loss}
Let us consider the image reconstruction error from a pair of images $I_{t^\prime}$ and $I_t$, distance estimate $\hat D_t$ at time $t$, and the relative pose for $I_t$, with respect to the source image $I_{t^\prime}$'s pose, as $T_{t \to t^\prime}$.
Using the distance estimate $\hat D_t$ of the network a point cloud $P_t$ is obtained via:
\begin{equation}
    \label{pt}
    P_t = \iPi(p_t, \hat D_t)
\end{equation}
where $\iPi$ represents the unprojection from image to camera coordinates as explained in Section~\ref{sec:img2cam}, $p_t$ the pixel set of image $I_t$. The pose estimate $T_{t \to t^\prime}$ from the pose network is used to get an estimate $\hat{P}_{t'} = T_{t \to t^\prime} P_t$ for the point cloud of the image $I_{t^\prime}$. $\hat{P}_{t^\prime}$ is then projected onto the fisheye camera at time $t'$ using the projection model $\Pi$ described in Section~\ref{sec:cam2img}. Combining transformation and projection with Eq.~\ref{pt} establishes a mapping from image coordinates $p_t=(u,v)^T$ at time $t$ to image coordinates $\hat p_{t^\prime}=(\hat u, \hat v)^T$ at time $t^\prime$. This mapping allows for the reconstruction $\hat I_{t' \to t}$ of the target frame $I_t$ by backward warping the source frame $I_{t^\prime}$.
\begin{align}
    \label{pixel_estimate}
    \hat{p}_{t'} = \Pi \big( T_{t \to t'} \iPi(p_t,\hat D_t) \big), \quad
    \hat{I}_{t' \to t}^{uv} = \Big\langle I_{t'}^{\hat{u}\hat{v}} \Big\rangle
\end{align}
Since the warped coordinates $\hat u, \hat v$ are continuous, we apply the differentiable spatial transformer network introduced by~\cite{jaderberg2015spatial} to compute $\hat{I}_{t' \to t}$ by performing bilinear interpolation of the four pixels from $I_{t^\prime}$ which lie close to $\hat p_{t'}$. The symbol $\big\langle\dots\big\rangle$ denotes the corresponding sampling operator.

Following~\cite{monodepth17, zhao2016loss} the image reconstruction error between the target image $I_t$ and the reconstructed target image $\hat I_{t' \to t}$ is calculated using the L1 pixel-wise loss term combined with Structural Similarity (SSIM)~\cite{wang2004image}, as our photometric loss $\mathcal{L}_{p}$ given by Eq.~\ref{eq:loss-photo} below.
\begin{align}
  \label{eq:loss-photo}
  \tilde{\mathcal{L}}_{p}(I_t,\hat I_{t' \to t}) &= \alpha~\frac{1 - \text{SSIM}(I_t,\hat I_{t' \to t}, \mathcal{M}_{t \to t^\prime})}{2} \nonumber \\
  &\quad+ (1-\alpha)~ \left\| (I_t - \hat I_{t' \to t}) \odot \mathcal{M}_{t \to t^\prime} \right\|_{l^1} \nonumber \\
    \mathcal{L}_{p} &= \min_{t^\prime \in \{t+1,t-1\}} \tilde{\mathcal{L}}_p(I_t, \hat I_{t' \to t})
\end{align}
where $\alpha = 0.85$, $\mathcal{M}_{t \to t^\prime}$ is the binary mask as discussed in Section~\ref{sec:mask} and the symbol $\odot$ denotes element-wise multiplication. Following~\cite{godard2019digging} instead of averaging the photometric error over all source images, we adopt per-pixel minimum. This significantly sharpens the occlusion boundaries and reduces the artifacts resulting in higher accuracy.

The self-supervised framework assumes a static scene, no occlusion and change of appearance (e.g. brightness constancy). A large photometric cost is incurred, potentially worsening the performance, if there exist dynamic objects and occluded regions. These areas are treated as outliers similar to \cite{zhou2018unsupervised} and clip the photometric loss values to a $95^\text{th}$ percentile. Zero gradient is obtained for errors larger than $95\%$. This improves the optimization process and provides a way to strengthen the photometric error.

\subsection{\textbf{\textit{Solving Scale Factor Ambiguity at Training Time}}}
\label{sec: scale-aware sfm}

For a pinhole projection model, $depth \propto 1 / disparity$. Henceforth, the network's sigmoided output $\sigma$ can be converted to depth with $D = 1 / ({a\sigma + b})$, where $a$ and $b$ are chosen to constrain $D$ between $0.1$ and $100$ units~\cite{godard2019digging}. For a spherical image, we can only obtain angular disparities~\cite{arican2009dense} by rectification. To perform distance estimation on raw fisheye images, we would require metric distance values to warp the source image $I_{t'}$ onto the target frame $I_t$. Due to the limitations of the monocular SfM objective, both the monocular depth $g_d$ and ego-motion predictor $g_\mathbf{x}$ predict \textit{scale-ambiguous} values which would make it impossible to estimate distance maps on fisheye images. To achieve scale-aware distance values, we normalize the pose network's estimate $T_{t \to t'}$ and scale it with $\Delta x$, the displacement magnitude relative to target frame $I_t$ which is calculated using vehicle's instantaneous velocity estimates $v_{t'}$ at time $t'$ and $v_t$ at time $t$. We also apply this technique on KITTI~\cite{geiger2013vision} to obtain metric depth maps.

\begin{equation}
    \overline{T}_{t \to t'} = \frac {T_{t \to t'}} {\|T_{t \to t'}\|} \cdot \Delta x
\end{equation}

\subsection{\textbf{\textit{Masking Static Pixels and Ego Mask}}} 
\label{sec:mask}

Following~\cite{godard2019digging}, we incorporate a masking approach to filter out static pixels which do not change their appearance from one frame to the other in the training sequence. The approach would filter out objects which move at the same speed as the ego-car, and also ignore the static frame when the ego-car stops moving. Similar to other approaches~\cite{godard2019digging, zhou2017unsupervised,vijayanarasimhan2017sfm,luo2019every} the per-pixel mask $\omega$ is applied to the loss by weighting the pixels selectively. Instead of being learned from the object motion~\cite{casser2019depth}, the mask is computed in the forward pass of the network, yielding a binary mask output where $\omega \in \{0, 1\}$. Wherever the photometric error of the warped image $\hat I_{t^\prime \to t}$ is not lower than that of the original unwarped source frame $I_{t^\prime}$ in each case compared to the target frame $I_t$, $\omega$ is set to ignore the loss of such pixels, \ie
\begin{align}
    \omega &= \big[ \, 
        \min_{t^\prime} pe(I_t, \hat I_{t^\prime \to t}) 
            < 
        \min_{t^\prime} pe(I_t, I_{t^\prime})     
            \, \big]
    \label{eqn:staticmask}
\end{align}
where $[\,]$ is the Iverson bracket. Additionally, we add a binary ego mask $\mathcal{M}_{t \to t^\prime}$ proposed in~\cite{mahjourian2018unsupervised} that ignores computing the photometric loss on the pixels that do not have a valid mapping \ie some pixel coordinates of the target image $I_t$ may not be projected onto the source image $I_{t'}$ given the estimated distance $\hat D_t$.

\subsection{\textbf{\textit{Backward Sequence}}}
\label{sec:backward sequence}

In the forward sequence, we synthesize the target frame $I_t$ with the source frames $I_{t-1}$ and $I_{t+1}$ (\ie as per above discussion $t' \in \{t+1, t-1\}$).
Analogously, backward sequence is carried out by using $I_{t-1}$ and $I_{t+1}$ as target frames and $I_t$ as source frame. We include warps $\hat I_{t \to t-1}$ and $\hat I_{t \to t+1}$, thereby inducing more constraints to avoid overfitting and resolve unknown distances in the border areas at the test time, as also observed in previous works~\cite{godard2019digging,zhou2017unsupervised,yin2018geonet}. We construct the loss for the additional backward sequence in a similar manner to the forward. This comes at the cost of high computational effort and longer training time as we perform two forward and backward warps which yields superior results on the Fisheye and KITTI dataset compared to the previous approaches~\cite{godard2019digging,zhou2017unsupervised} which train only with one forward sequence and one backward sequence.

\subsection{\textbf{\textit{Edge-Aware Smoothness Loss}}}

In order to regularize distance and avoid divergent values in occluded or texture-less low-image gradient areas, we add a geometric smoothing loss. We adopt the edge-aware term similar to~\cite{monodepth17, mahjourian2018unsupervised, zou2018df}. The regularization term is imposed on the inverse distance map. Unlike previous works, the loss is not decayed for each pyramid level by a factor of $2$ due to down-sampling, as we use a super resolution network (see Section~\ref{sec:deformable super-resolution network})
\begin{equation}
    \mathcal{L}_{s}(\hat{D}_t) = | \partial_u \hat{D}^*_t | e^{-|\partial_u I_t|} + | \partial_v \hat{D}^*_t | e^{-|\partial_v I_t|}
\end{equation}
To discourage shrinking of estimated distance~\cite{Wang_2018_CVPR}, mean-normalized inverse distance of $I_t$ is considered, i.e. $\hat{D}^*_t = \hat{D}^{-1}_t / \overline{D}_t$, where $\overline{D}_t$ denotes the mean of $\hat{D}^{-1}_t := 1 /\hat{D}_t$.

\subsection{\textbf{\textit{Cross-Sequence Distance Consistency Loss}}}

The SfM setting uses an N-frame training snippet $S = \{ {I_1},{I_2}, \cdots ,{I_N}\} $ from a video as input. The FisheyeDistanceNet can estimate the distance of each image in the training sequence. Another constraint can be enforced among the frames in $S$, since the distances of a 3D point estimated from different frames should be consistent.

Let us assume $\hat D_{t'}$ and $\hat D_{t}$ are the estimates of the images $I_{t'}$ and $I_t$ respectively. For each pixel ${{p}_{t}}\in {{I}_{t}}$, we can use Eq.~\ref{pixel_estimate} to obtain $\hat{p}_{t'}$. Since it's coordinates are real valued, we apply the differentiable spatial transformer network introduced by~\cite{jaderberg2015spatial} and estimate the distance value of $\hat{p}_{t'}$ by performing bilinear interpolation of the four pixel's values in $\hat D_{t'}$ which lie close to $\hat p_{t'}$. Let us denote the distance map obtained through this as ${{\hat{D}}_{t \to t'}}\left(p_{t}\right)$. 
Next, we can transform the point cloud in frame $I_t$ to frame $I_{t'}$ by first obtaining $P_t$ using Eq. ~\ref{pt}. We transform the point cloud $P_t$ using the pose network's estimate via $\hat{P}_{t'} = T_{t \to t'} P_t$. Now, ${{D}_{t \to t'}}\left({{p}_{t}}\right) := \| \hat{P}_{t'} \|$ denotes the distance generated from point cloud $\hat{P}_{t'}$.
Ideally, ${{D}_{t \to t'}}\left( {{p}_{t}} \right)$ and ${{\hat{D}}_{t \to t'}}\left( p_{t} \right)$ should be equal. Therefore, we can define the following cross-sequence distance consistency loss (CSDCL) for the training sequence~$S$:
\begin{align} 
    \label{equ:dclf}
	\mathcal{L}_{dc} = \sum_{t=1}^{N-1} \sum_{t'=t+1}^{N} \bigg( &\sum_{p_t} \mathcal{M}_{t \to t^\prime}
	\left| D_{t \to t^\prime}\left(p_t \right) - \hat D_{t \to t^\prime}\left(p_t \right) \right| \nonumber \\
	+ &\sum_{p_{t'}} \mathcal{M}_{t' \to t} \left| D_{t' \to t}\left(p_{t'} \right) - \hat D_{t' \to t}\left(p_{t'} \right) \right| \bigg)
\end{align}
Eq. \ref{equ:dclf} contains one term for which pixels and point clouds are warped forwards in time (from $t$ to $t'$) and one term for which they are warped backwards in time (from $t'$ to $t$).

In prior works~\cite{vijayanarasimhan2017sfm,zou2018df}, the consistency error is limited to only two frames, whereas we apply it to the entire training sequence $S$. This induces more constraints and enlarges the baseline, inherently improving the distance estimation~\cite{zhou2018unsupervised}.

\subsection{\textbf{\textit{Final Training Loss}}}
The overall self-supervised structure-from-motion (SfM) \textit{from motion} objective consists of a photometric loss $\mathcal{L}_p$  imposed between the reconstructed target image $\hat{I}_{t' \to t}$ and the target image $I_t$, included once for the forward and once for the backward sequence, and a distance regularization term $\mathcal{L}_s$ ensuring edge-aware smoothing in the distance estimates. Finally, $\mathcal{L}_{dc}$ a cross-sequence distance consistency loss derived from the chain of frames in the training sequence $S$ is also included. To prevent the training objective getting stuck in the local minima due to the gradient locality of the bilinear sampler~\cite{jaderberg2015spatial}, we adopt 4 scales to train the network as followed in~\cite{zhou2017unsupervised,monodepth17}. The final objective function is averaged over per-pixel, scale and image batch.
\begin{align} 
    \label{equ:objective}
    \mathcal{L} &= \sum\limits_{n = 1}^4 {\frac{{\mathcal{L}_{n}}}{{{2^{n - 1}}}}} ,\\
    \mathcal{L}_{n} &= {}^n\mathcal{L}_{p}^f + {}^n\mathcal{L}_{p}^b + \gamma\ {}^n{\mathcal{L}_{dc}} + \beta\ {}^n{\mathcal{L}_{s}} \nonumber
\end{align}
\begin{table*}[!ht]
\vspace{2mm}
\captionsetup{singlelinecheck=false, font=small, labelsep=space, belowskip=-2pt}
\centering
\begin{tabular}{lccccccc}
    \toprule
    & \cellcolor[HTML]{5880ab}Abs Rel & \cellcolor[HTML]{5880ab}Sq Rel & \cellcolor[HTML]{5880ab}RMSE &  \cellcolor[HTML]{5880ab}RMSE$_{log}$ & \cellcolor[HTML]{e8715b}$\delta < 1.25$ & \cellcolor[HTML]{e8715b}$\delta < 1.25^2$ & \cellcolor[HTML]{e8715b}$\delta < 1.25^3$ \\
    \cmidrule(lr){2-5} \cmidrule(lr){6-8}
    \textbf{Approach} & \multicolumn{4}{c}{\cellcolor[HTML]{5880ab}lower is better} & \multicolumn{3}{c}{\cellcolor[HTML]{e8715b}higher is better}\\
    \midrule
    \multicolumn{8}{c}{\cellcolor[HTML]{448BE9}\textbf{\textit{KITTI}}} \\
    \midrule
    Zhou~\cite{zhou2017unsupervised}\textdagger       & 0.183 & 1.595 & 6.709 & 0.270 & 0.734 & 0.902 & 0.959 \\
    Yang~\cite{yang2018unsupervised}                  & 0.182 & 1.481 & 6.501 & 0.267 & 0.725 & 0.906 & 0.963 \\
    Vid2depth~\cite{mahjourian2018unsupervised}       & 0.163 & 1.240 & 6.220 & 0.250 & 0.762 & 0.916 & 0.968 \\
    GeoNet~\cite{yin2018geonet}\textdagger            & 0.149 & 1.060 & 5.567 & 0.226 & 0.796 & 0.935 & 0.975 \\
    DDVO~\cite{Wang_2018_CVPR}                        & 0.151 & 1.257 & 5.583 & 0.228 & 0.810 & 0.936 & 0.974 \\
    DF-Net~\cite{zou2018df}                           & 0.150 & 1.124 & 5.507 & 0.223 & 0.806 & 0.933 & 0.973 \\
    Ranjan~\cite{ranjan2019competitive}               & 0.148 & 1.149 & 5.464 & 0.226 & 0.815 & 0.935 & 0.973 \\
    EPC++~\cite{luo2019every}                         & 0.141 & 1.029 & 5.350 & 0.216 & 0.816 & 0.941 & 0.976 \\
    Struct2depth `(M)'~\cite{casser2019depth}         & 0.141 & 1.026 & 5.291 & 0.215 & 0.816 & 0.945 & 0.979 \\
    Zhou~\cite{zhou2018unsupervised}                  & 0.139 & 1.057 & 5.213 & 0.214 & 0.831 & 0.940 & 0.975 \\
    PackNet-SfM~\cite{guizilini2019packnet}           & 0.120 & 0.892 & 4.898 & 0.196 & 0.864 & 0.954 & 0.980 \\
    Monodepth2~\cite{godard2019digging}               & \textbf{0.115} & 0.903 & 4.863 & 0.193 & \textbf{0.877} & 0.959 & 0.981 \\
    \textbf{FisheyeDistanceNet}                       & 0.117 & \textbf{0.867} & \textbf{4.739} & \textbf{0.190} & 0.869 & \textbf{0.960} & \textbf{0.982} \\
    \textbf{FisheyeDistanceNet} (1024 $\times$ 320)   & 0.109 & 0.788 & 4.669 & 0.185 & 0.889 & 0.964 & 0.982 \\
    \midrule
    \multicolumn{8}{c}{\cellcolor[HTML]{34FF34}\textbf{\textit{WoodScape}}} \\
    \midrule
    FisheyeDistanceNet cap $80\,\text{m}$             & 0.167 & 1.108 & 3.814 & 0.216 & 0.794 & 0.953 & 0.972 \\
    FisheyeDistanceNet cap $40\,\text{m}$             & 0.152 & 0.768 & 2.723 & 0.210 & 0.812 & 0.954 & 0.974 \\
    FisheyeDistanceNet cap $30\,\text{m}$             & 0.149 & 0.613 & 2.402 & 0.204 & 0.810 & 0.957 & 0.976 \\
    \bottomrule
\end{tabular}
\caption{\textbf{Quantitative results of leaderboard algorithms on KITTI dataset~\cite{geiger2013vision} and FisheyeDistanceNet on Fisheye dataset part of WoodScape~\cite{yogamani2019woodscape}}. Single-view depth estimation results using the Eigen Split~\cite{Eigen_14} for depths reported less than $80\,m$, as indicated in~\cite{Eigen_14} for pinhole model. All the approaches are self-supervised on monocular video sequences. At test-time, all monocular methods excluding our FisheyeDistanceNet, scale the estimated depths using median ground-truth LiDAR depth. For the fisheye dataset, we estimate distance rather than depth. \textdagger~marks newer results reported on GitHub.}
\label{tab:results}
\end{table*}
\section{\textbf{Network Details}}
\subsection{\textbf{\textit{Deformable Super-Resolution Distance and PoseNet}}}
\label{sec:deformable super-resolution network}

The distance estimation network is mainly based on the U-net architecture~\cite{ronneberger2015u}, an \textit{encoder-decoder} network with skip connections. After testing different ResNet family variants, such as ResNet50 with 25M parameters, we chose a ResNet18~\cite{he2016deep} as the encoder. The key aspect here is replacing normal convolutions with deformable convolutions since regular CNNs are inherently limited in modeling large, unknown geometric distortions due to their fixed structures, such as fixed filter kernels, fixed receptive field sizes, and fixed pooling kernels~\cite{dai2017deformable,zhu2019deformable}.

In previous works~\cite{godard2019digging,zhou2017unsupervised,monodepth17,Wang_2018_CVPR,yin2018geonet}, the decoded features were upsampled via a nearest-neighbor interpolation or with learnable transposed convolutions. This process's main drawback is that it may lead to large errors at object boundaries in the upsampled distance map as the interpolation combines distance values of background and foreground. For effective and detailed preservation of the decoded features, we leverage the concept of sub-pixel convolutions~\cite{shi2016real} to our super-resolution network. We use pixel shuffle convolutions and replace the convolutional feature upsampling, performed via a nearest-neighbor interpolation or with learnable transposed convolutions. The resulting distance maps are super-resolved, have sharp boundaries, and expose more details of the scene.

The backbone of our pose estimation network is based on~\cite{godard2019digging} and predicts rotation using Euler angle parameterization.
The output is a set of six DOF transformations between $I_{t-1}$ and $I_t$ as well as $I_{t}$ and $I_{t+1}$. We have replaced normal convolutions with deformable convolutions for the encoder-decoder setting.

\subsection{\textbf{Implementation Details}}

We use Pytorch~\cite{paszke2017automatic} and employ Adam~\cite{kingma2014adam} optimizer to minimize the training objective function (\ref{equ:objective}) with ${{\beta }_{1}}=0.9$, ${{\beta }_{2}}=0.999$. We train the model for 25 epochs, with a batch size of 20 on 24GB Titan RTX with initial learning rate of ${{10}^{-4}}$ for the first 20 epochs, then drop to ${{10}^{-5}}$ for the last 5 epochs. The sigmoided output $\sigma$ from the distance decoder
is converted to distance with  $D = {a \cdot \sigma + b}$. For the pinhole model, depth $D =  1 / ({a \cdot \sigma + b})$, where $a$ and $b$ are chosen to constrain $D$ between $0.1$ and $100$ units. The original input resolution of the fisheye image is $1280\times800$ pixels, we crop it to $1024\times512$ to remove the vehicle's bumper, shadow and other artifacts of the vehicle. Finally the cropped image is downscaled to $512\times256$ before feeding to the network.
For the pinhole model on KITTI, we use $640\times192$ pixels as the network input.

We experimented with batch normalization~\cite{ioffe2015batch} and group normalization~\cite{wu2018group} layers in the 
encoder-decoder setting. We have found that group normalization with $G=32$ significantly improves the results~\cite{he2019rethinking}.
The smoothness weight term $\beta$ and cross-sequence distance consistency weight term $\gamma$ have been set to $0.001$. We applied deformable convolutions to the 3 x 3 Conv layers in stages conv3, conv4, and conv5 in ResNet18 and ResNet50, with 12 layers of deformable convolution in the encoder part compared to 3 layers in \cite{dai2017deformable}, all in the conv5 stage for ResNet50. We replaced the subsequent layers of the decoder with deformable convolutions for the distance and pose network. For the pinhole model, on KITTI Eigen split in Section~\ref{sec:kitti_eigen_split}, we used normal convolutions instead of deformable convolutions. Finally, to alleviate checkerboard artifacts from the output distance maps using sub-pixel convolution~\cite{shi2016real}, we initialized the last convolutional layer in a specific way before the pixel shuffle operation as described in~\cite{aitken2017checkerboard}.
\begin{figure*}[!ht]
\vspace{0.15cm}
  \captionsetup{singlelinecheck=false, font=footnotesize, labelsep=space}
  \centering
  \resizebox{\textwidth}{!}{
  \newcommand{\turnheightnew}{0.25\columnwidth}
\centering

\begin{tabular}{@{\hskip 0.5mm}c@{\hskip 0.5mm}c@{\hskip 0.5mm}c@{\hskip 0.5mm}c@{\hskip 0.5mm}c@{}}

{\rotatebox{90}{\hspace{6mm}Raw Input}} &
\includegraphics[height=\turnheightnew]{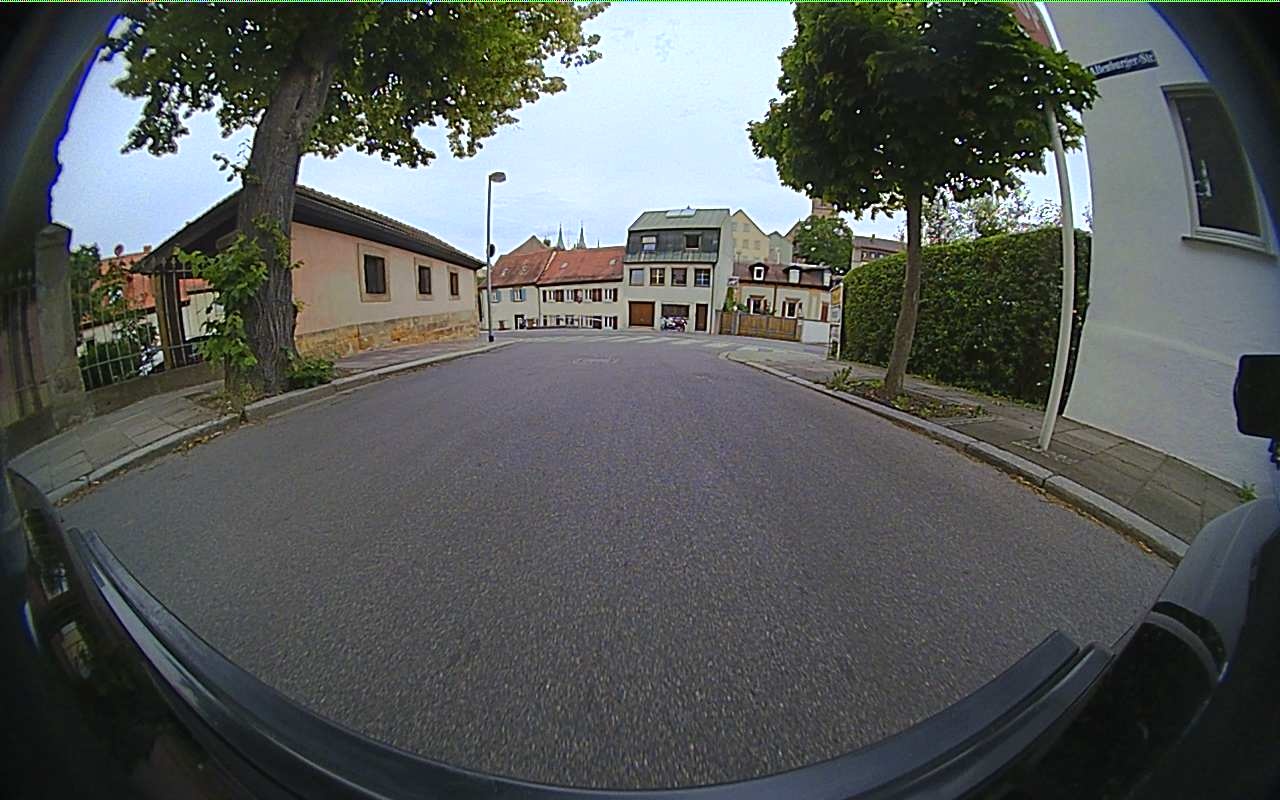} &
\includegraphics[height=\turnheightnew]{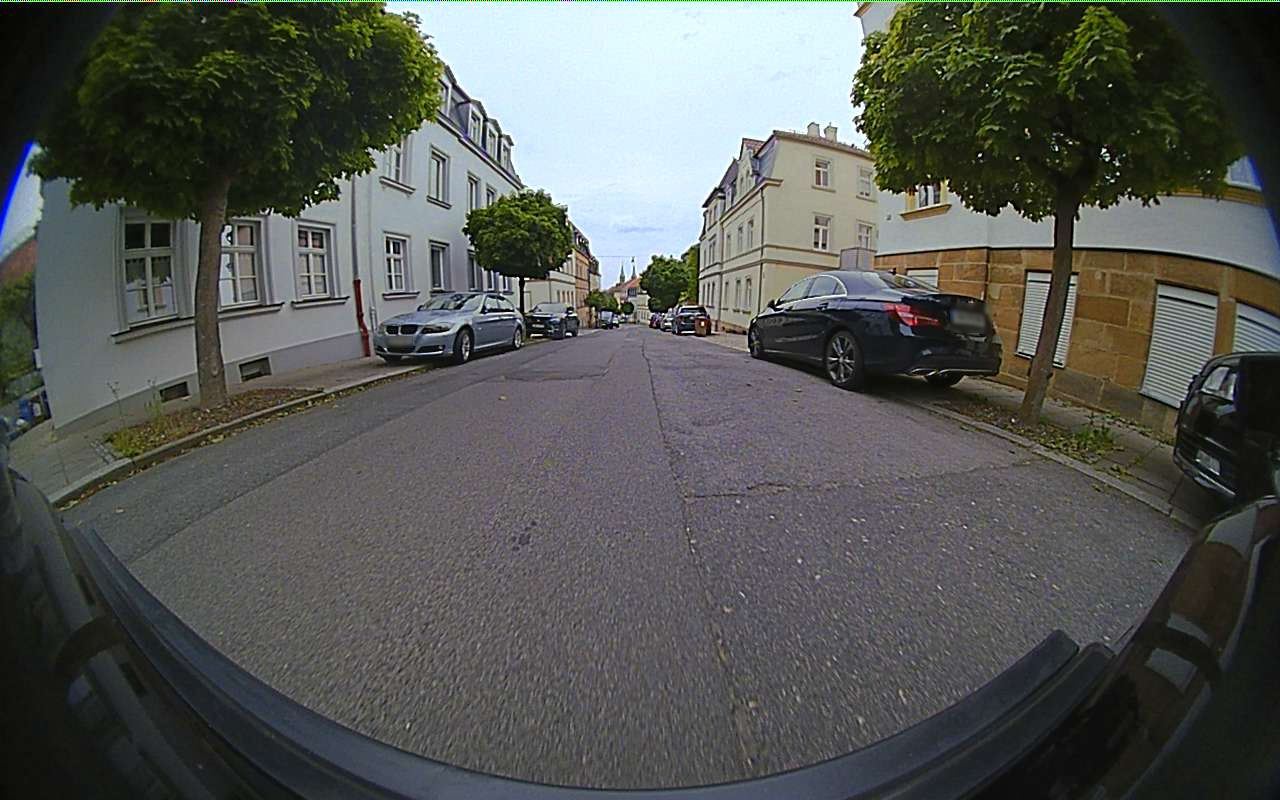} &
\includegraphics[height=\turnheightnew]{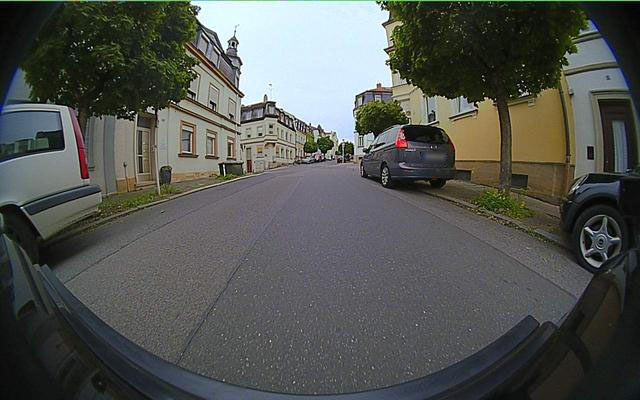} &
\includegraphics[height=\turnheightnew]{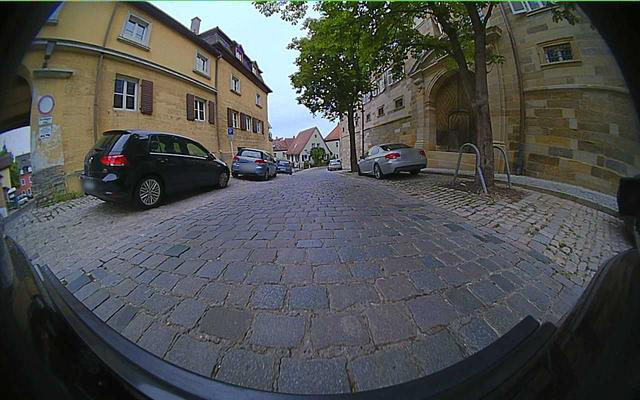}\\

{\rotatebox{90}{\hspace{0mm}Cropped Input}} &
\includegraphics[height=\turnheightnew]{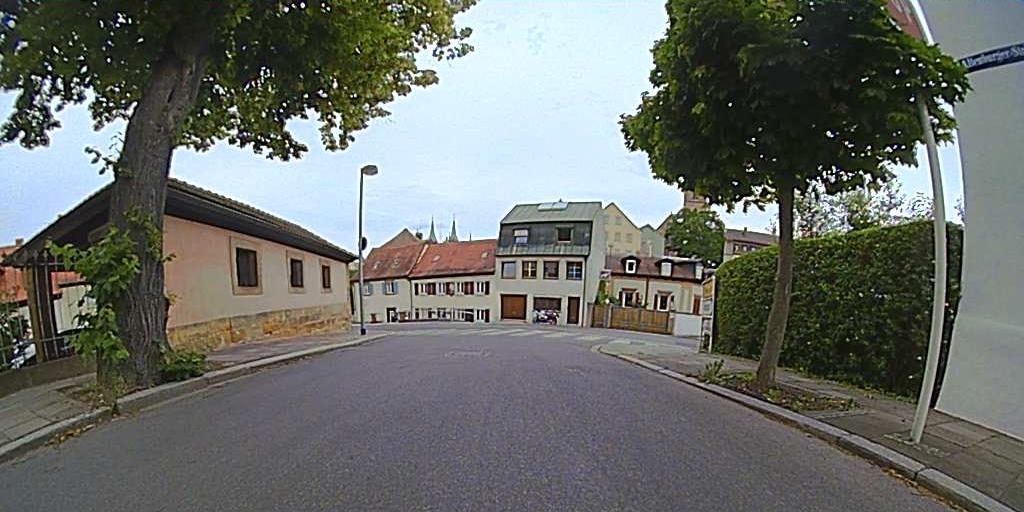} &
\includegraphics[height=\turnheightnew]{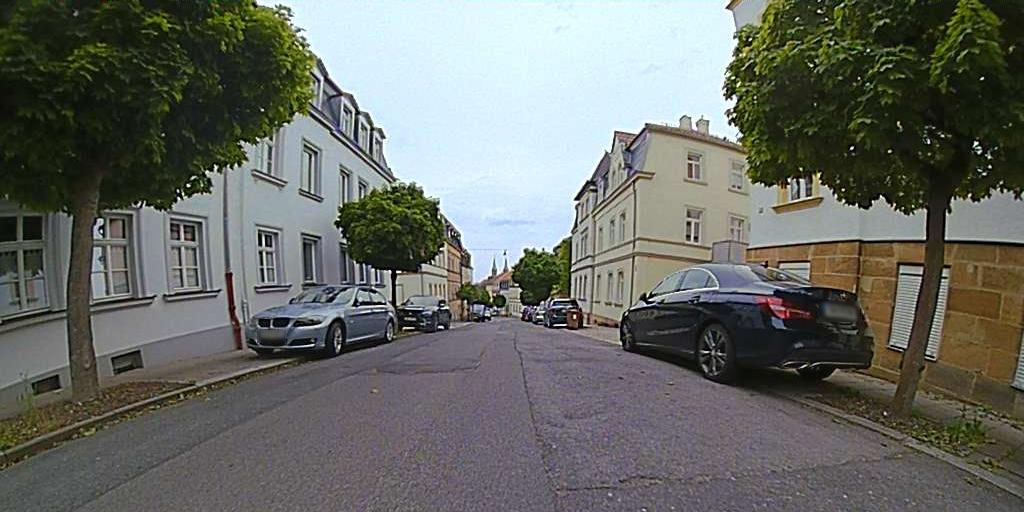} &
\includegraphics[height=\turnheightnew]{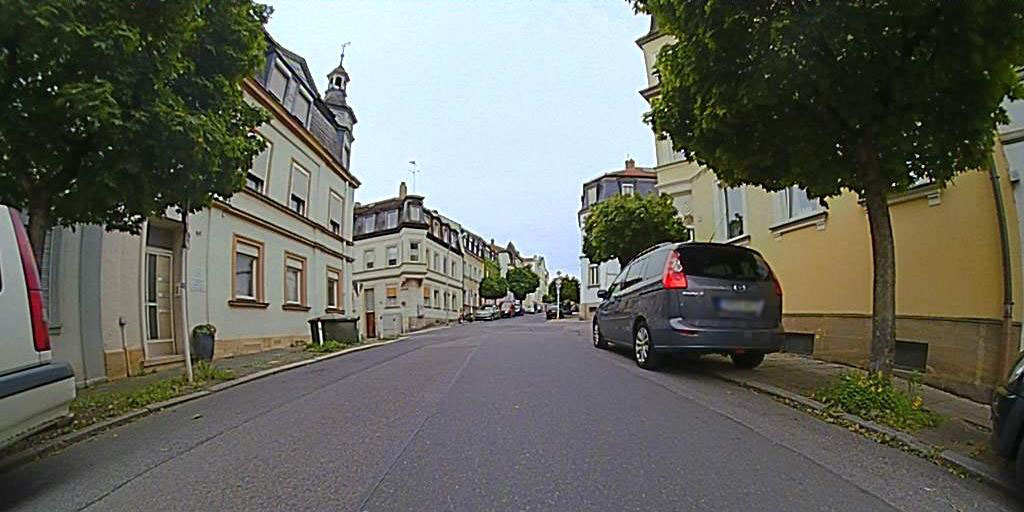} &
\includegraphics[height=\turnheightnew]{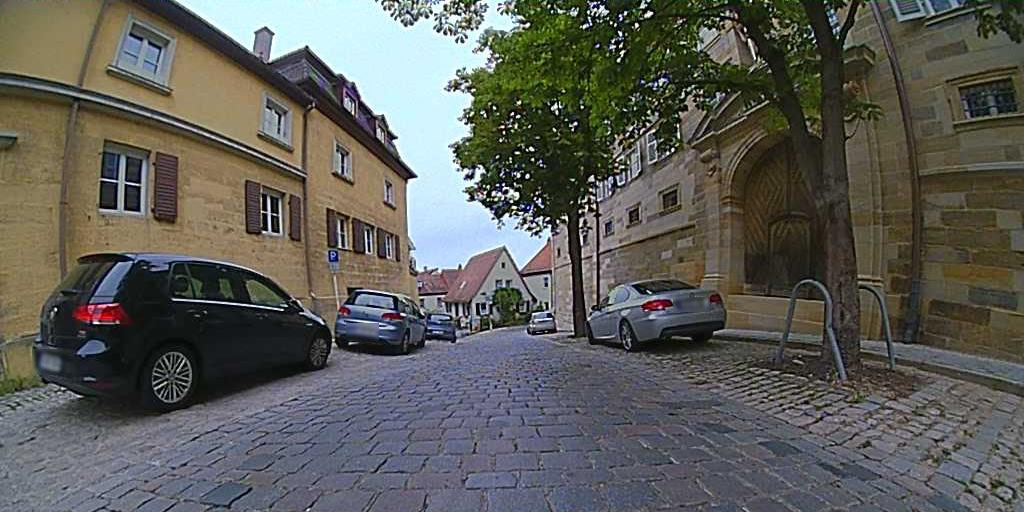}\\

{\rotatebox{90}{\hspace{0mm}\scriptsize}} &
\includegraphics[height=\turnheightnew]{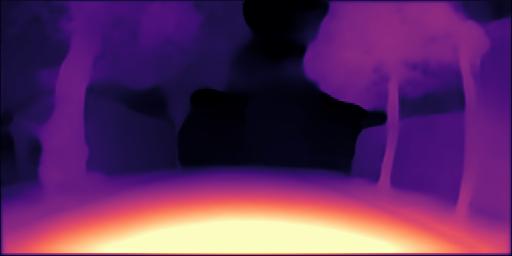} &
\includegraphics[height=\turnheightnew]{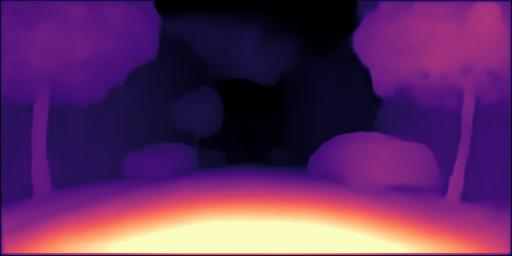} &
\includegraphics[height=\turnheightnew]{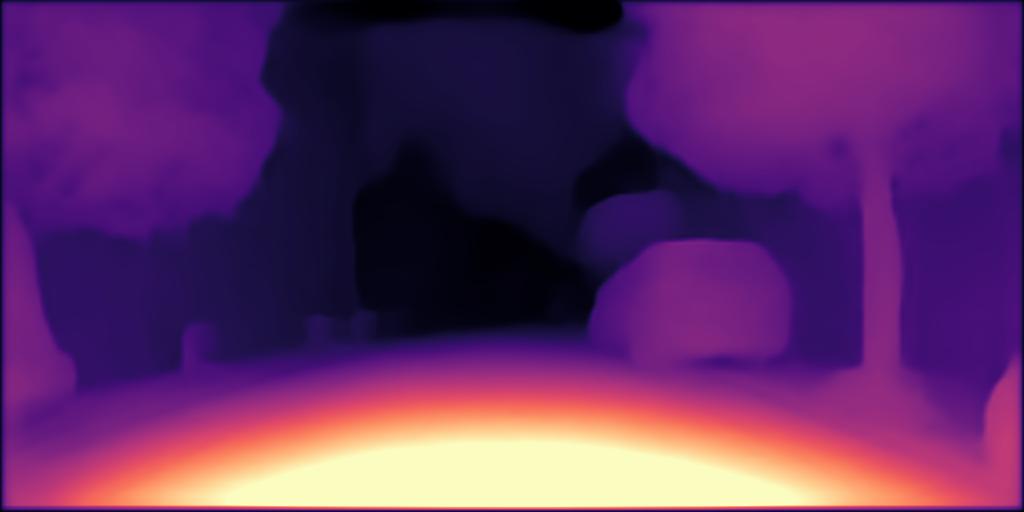} &
\includegraphics[height=\turnheightnew]{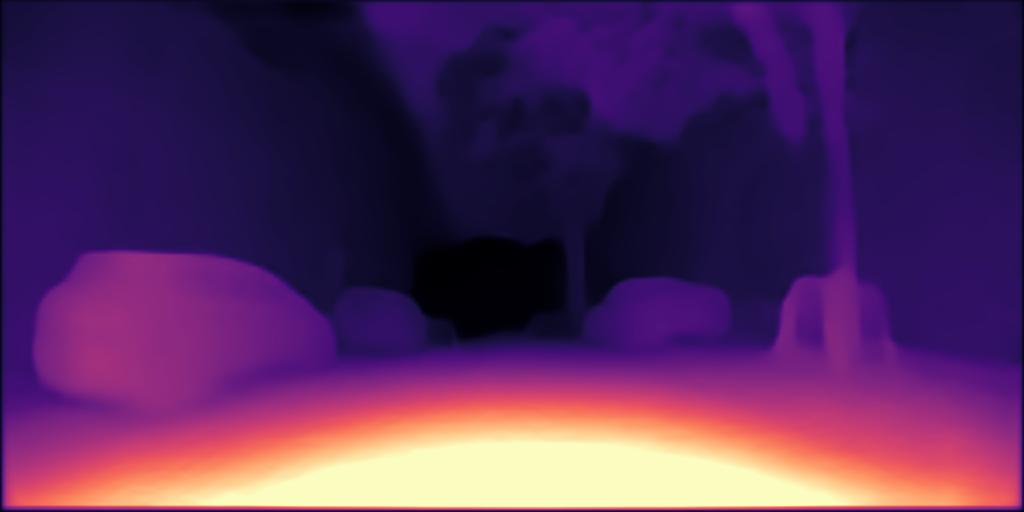} \\

\end{tabular}
}
  \caption{\textbf{\textbf{Qualitative results on the Fisheye WoodScape dataset.}}
  Our FisheyeDistanceNet produces sharp distance maps on raw fisheye images.}
  \label{fig:fisheye_qual}
\end{figure*}

\begin{table*}[t!]
    \captionsetup{singlelinecheck=false, font=small, labelsep=space, belowskip=-14pt}
	\small
	\begin{center}
	\begin{tabular}{l|c|c|c|c|c|c|c|c|c|c|c|c}
	\toprule
	Method & FS & BS & SR & CSDCL & DCN & \cellcolor[HTML]{5880ab}Abs Rel & \cellcolor[HTML]{5880ab}Sq Rel & \cellcolor[HTML]{5880ab}RMSE & \cellcolor[HTML]{5880ab}RMSE$_{log}$ & \cellcolor[HTML]{e8715b}$\delta < 1.25$ & \cellcolor[HTML]{e8715b}$\delta < 1.25^2$ &  \cellcolor[HTML]{e8715b}$\delta < 1.25^3$\\
	\toprule
	Ours & \ch & \ch & \ch & \ch & \ch & 0.152 & 0.768 & 2.723 & 0.210 & 0.812 & 0.954 & 0.974 \\
	Ours & \ch & \xm & \ch & \ch & \ch & 0.172 & 0.829 & 2.925 & 0.243 & 0.802 & 0.952 & 0.970 \\
	Ours & \ch & \xm & \xm & \ch & \ch & 0.181 & 0.913 & 3.180 & 0.250 & 0.823 & 0.938 & 0.963 \\
	Ours & \ch & \xm & \xm & \xm & \ch & 0.190 & 0.997 & 3.266 & 0.258 & 0.796 & 0.930 & 0.963 \\
    Ours & \ch & \xm & \xm & \xm & \xm & 0.201 & 1.282 & 3.589 & 0.276 & 0.590 & 0.898 & 0.949 \\
	\bottomrule
\end{tabular}
\end{center}
\caption{\textbf{Ablation study on different variants of our FisheyeDistanceNet using the Fisheye WoodScape dataset~\cite{yogamani2019woodscape}}. Distances are capped at $40\,m$. BS, SR, CSDCL, and DCN represent a backward sequence, super-resolution network with PixelShuffle, or sub-pixel convolution initialized to convolution NN resize (ICNR)~\cite{aitken2017checkerboard}, cross-sequence distance consistency loss, and deformable convolutions respectively. The input resolution is $512 \times 256$ pixels.}
\label{table:ablation}
\vspace{-1mm}
\end{table*}

\section{\textbf{Experiments}}
\subsection{\textbf{\textit{Datasets}}}
\textbf{\textit{WoodScape -- Fisheye Dataset}}
The dataset contains roughly 40,000 raw images obtained with a fisheye camera and point clouds from a sparse Velodyne HDL-64E rotating 3D laser scanner as ground truth for the test set. The training set contains 39,038 images collected by driving around various parts of Bavaria, Germany. The validation and the test split contain 1,214 and 697 images, respectively. The dataset distribution is similar to the KITTI Eigen split used in~\cite{godard2019digging,zhou2017unsupervised} for the pinhole model. The training set comprises three scene categories: \textit{city}, \textit{residential} and \textit{sub-urban}. While training, these categories are randomly shuffled and fed to the network. We filter static scenes based on the speed of the vehicle with a threshold of $2\,\text{km}/\text{h}$ to remove image frames that only observe minimal camera ego-motion, since distance cannot be learned under these circumstances. Comparable to previous experiments on pinhole SfM~\cite{zhou2017unsupervised, godard2019digging}, we set the length of the training sequence to 3.

\textbf{\textit{KITTI -- Eigen Split}}
\label{sec:kitti_eigen_split}
We use the KITTI dataset, and data split according to Eigen et al.~\cite{eigen2015predicting} for the experiments with pinhole image data. We filter static frames as proposed by Zhou et al.~\cite{zhou2017unsupervised}. The resulting training set contains 39,810 images, and the validation split comprises 4,424 images. We use the standard test set of 697 images. The length of the training sequence is set to 3.

\subsection{\textbf{Evaluation}}

We evaluate FisheyeDistanceNet's depth and distance estimation results using the metrics proposed by Eigen et al.~\cite{Eigen_14} to facilitate comparison. The quantitative results are shown in the Table~\ref{tab:results} illustrate that our scale-aware self-supervised approach outperforms all the state-of-the-art monocular approaches. We could not leverage the Cityscapes dataset into our training regime to benchmark our scale-aware framework due to the absence of odometry data.

Since the projection operators are different, previous SfM approaches will not be feasible on the Fisheye Woodscape dataset without adapting the network and projection model. It is important to note that due to the fisheye's geometry, it would not be a fair comparison to evaluate the distance estimates up to $80\,m$. Our fisheye automotive cameras also undergo high data compression, and our dataset contains images of inferior quality compared with KITTI. Our fisheye cameras can perform well up to a range of $40\,m$. Therefore, we also report results on a $30\,m$, and a $40\,m$. range (see Table~\ref{tab:results}).
\subsection{\textbf{Fisheye Ablation Study}}
We conduct an ablation study to evaluate the importance of different components. We ablate the following components and report their impact on the distance evaluation metrics in Table~\ref{table:ablation}: (i) \textit{Remove Backward Sequence}: The network is only trained for the forward sequence which consists of two warps as explained in Section~\ref{sec:backward sequence}; (ii) \textit{Additionally remove Super-Resolution using sub-pixel convolution}: Removal of sub-pixel convolution has a huge impact on Woodscape compared to KITTI. This is mainly attributed to the fisheye model, as far-away objects are tiny and cannot be resolved accurately with naive nearest-neighbor interpolation or transposed convolution~\cite{odena2016deconvolution}; (iii) \textit{Additionally remove cross-sequence distance consistency loss}: Removing the CSDCL mainly diminishes the baseline; (iv) \textit{Additionally remove deformable convolutions}: If we remove all the major components, especially deformable convolution layers~\cite{zhu2019deformable}, our model will fail miserably as the distortion introduced by the fisheye model will not be learned correctly by normal convolutional layers.
\section{\textbf{Conclusion}}
\vspace{-0.05cm}
We propose a novel self-supervised training strategy to obtain metric distance maps on unrectified fisheye images. Through extensive experiments, we show that our FisheyeDistanceNet establishes a new state-of-the-art in the self-supervised monocular distance and depth estimation on Fisheye WoodScape and KITTI dataset, respectively. We obtain promising results demonstrating the potential of using a CNN based approach for deployment in commercial, automotive systems, particularly for replacing current classical depth estimation approaches. To encourage further research on fisheye distance estimation, we will release the dataset as a part of WoodScape \cite{yogamani2019woodscape} project.
\clearpage
\bibliographystyle{IEEEtran}
\bibliography{bib/ieee}
\clearpage
\end{document}